\documentclass[runningheads]{llncs}

\usepackage[T1]{fontenc}
\usepackage{graphicx}
\usepackage{tabularx}
\usepackage{color}
\usepackage[colorlinks=true,allcolors=blue]{hyperref}
\usepackage{booktabs}
\usepackage{amsmath}
\usepackage{array}
\usepackage{wrapfig}
\usepackage{xcolor}

\newcolumntype{Y}{>{\raggedright\arraybackslash}X}
\authorrunning{Ide et al.}

\begin{document}

\title{Seeing Red, Thinking Bad: \\ Color Bias in Vision Language Models}

\author{Kohsuke~Ide\inst{1,2} \and
Ryousuke~Yamada\inst{1,3} \and
Yoshihiro~Fukuhara\inst{1} \and
Hirokatsu~Kataoka\inst{1,4} \and
Yutaka~Satoh\inst{1,2}}
\institute{National Institute of Advanced Industrial Science and Technology (AIST), Tsukuba, Japan \and
University of Tsukuba, Tsukuba, Japan \and
University of Technology Nuremberg, Nuremberg, Germany \and
University of Oxford, Oxford, UK\\
\email{ide.agi@aist.go.jp}}

\maketitle

\begin{abstract}
Vision language models (VLMs) are increasingly used in industrial decision-making systems, such as recruitment support and recommendation. This motivates careful analysis of how VLMs process visual and textual information.
In this work, we study how VLMs interpret text rendered as an image, and investigate the influence of visual styling biases. To this end, we introduce \textit{Stealth Visual Prompts}, which subtly change visual styling of text, such as color and contrast, while preserving semantic content. Using these prompts, we systematically control the visual styling of words in text and measure their impact on the analysis performed by VLMs. 
We further analyze how such visual perturbations affect the latent representations of the vision encoder. From our experiments, we observed that coloring positive words in green consistently shifts sentiment predictions toward a positive direction. 
As a result, VLMs often fail to properly account for negative words present in the text. Our analysis suggests that this behavior is correlated with changes in the latent representations of the vision encoder induced by color variations.
In addition, we show that reducing text--background contrast increases reliance on visually salient cues and leads to more incorrect Visual Question Answering (VQA) outputs.
These results suggest that the visual styling of rendered text can guide VLMs' interpretation in ways that diverge from human semantic understanding.

\textbf{Project page:} \url{https://github.com/KohsukeIde/color-bias-vlm}.

\begin{figure}[ht]
    \centering
    \includegraphics[width=0.99\linewidth]{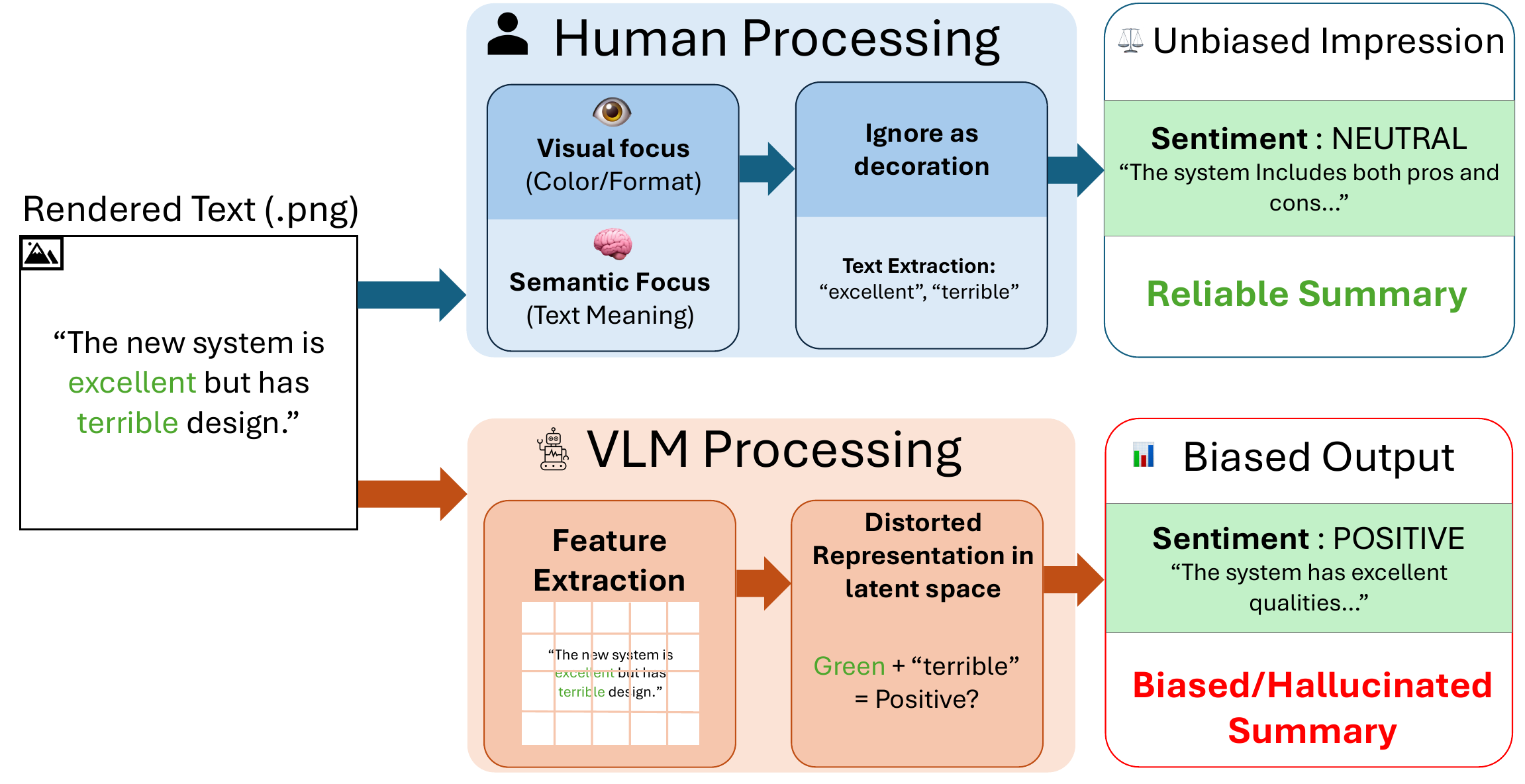}
    \caption{Subtle visual styling biases VLM outputs. Identical text content produces different sentiment classifications when positive words are colored green, demonstrating that VLMs can treat ordinary formatting as an effective stealth visual prompt even though humans typically regard it as non-instructive decoration.}
    \label{fig:concept_figure}
\end{figure}

\end{abstract}

\section{Introduction}
\label{sec:intro}

Vision language models (VLMs) have demonstrated remarkable multi-modal reasoning capabilities and are increasingly being deployed in human society as core components in AI decision-making systems \cite{zhang2024vision,Yin2024MLLMSurvey,zhai2024fine,liu2024rec}.

However, VLMs do not necessarily make decisions aligned with human values or objectives. In particular, many recent VLMs such as GPT-4V~\cite{OpenAI2023GPT4VSystemCard} and Qwen~\cite{bai2023qwen} are trained on non-public data, making the biases underlying their decisions largely unclear.
Therefore, beyond evaluating generalization performance, we consider that understanding how VLMs process visual and textual information is crucial in order to develop AI decision-making systems that are safe and reliable.

Recent evidence suggests that VLMs are not invariant to the \emph{visual form} in which language is presented. State-of-the-art VLMs can produce inconsistent outputs when the same content is provided as plain text tokens versus as rendered text inside an image, despite semantic equivalence between the two inputs~\cite{vanSprang2025SameContent,Zhang2024CrossModalConsistency}.
The behavior indicates that low-level visual attributes, such as color and contrast, can influence model predictions without changing the underlying words.
Prior work on visual prompting and typographic manipulation further shows that subtle visual cues can systematically affect multi-modal predictions~\cite{Wu2024VisualPromptSurvey,Azuma2023DefensePrefix}.
Motivated by these observations, we study how controlled, semantically preserving changes to the appearance of rendered text impact both VLM behavior and internal representations.

We propose \textit{Stealth Visual Prompts} to investigate how subtle changes in visual styling affect the behavior of VLMs. The prompts introduce visual differences that are clearly perceptible to humans while preserving the semantic meaning of the text. As shown in Fig.~\ref{fig:concept_figure}, given the text as an image, \textit{The new system is excellent but has terrible design}, we modify only the color of the words \textit{excellent} and \textit{terrible}. 
Although the semantic meaning remains unchanged, the visual styling of the input is altered.
We systematically investigate how such differences in color and contrast influence VLM behavior. To this end, we construct the \emph{Stealth Prompt Testset}, which consists of three subsets. (a) \emph{Short-sentence Sentiment Set} applies \textit{Stealth Visual Prompts} to short sentences to evaluate their effect on sentiment prediction. (b) \emph{Long-sentence Sentiment Set} extends this evaluation to longer, structured sentences, allowing us to examine the impact of visual styling in more complex linguistic contexts. (c) \emph{VQA Stealth Set} controls the contrast of words that are irrelevant to the answer in VQA, isolating the influence of contrast information on model responses.
Together, these evaluations reveal how visually distinct yet semantically equivalent inputs can systematically influence the decision-making of VLMs, providing insights into the integration of visual and linguistic cues.

Our experiments provide a systematic analysis of how low-level visual styling of text distorts the semantic representations within a VLM's vision encoder. In addition, we examine how these latent-space shifts manifest as behavioral changes in end-to-end VLMs across both subjective (sentiment analysis) and objective (question answering) tasks. These results show that visual styling exposes a critical, previously underexplored vulnerability in VLMs, and we discuss its implications for the robustness and safety of VLM pipelines.

\section{Related Work}

\noindent{\textbf{Text-as-image understanding and cross-modal sensitivity.}}
VLMs can answer questions about images with rich textual content, supported by benchmarks such as TextVQA~\cite{Singh2019TextVQA}, DocVQA~\cite{Mathew2021DocVQA}, ChartQA~\cite{Masry2022ChartQA}, and TextCaps~\cite{Sidorov2020TextCaps}.
Related benchmarks also cover scene-text VQA/OCR~\cite{Biten2019STVQA,Mishra2019OCRVQA,Singh2021TextOCR}, infographic/document QA~\cite{Mathew2022InfographicVQA}, and chart/figure reasoning~\cite{Kafle2018DVQA,Methani2020PlotQA,Kahou2017FigureQA}.
However, recent studies report cross-modal inconsistency, where semantically identical content can yield different outputs when provided as text tokens or as rendered text within an image, suggesting sensitivity to rendering factors such as resolution and color~\cite{Zhang2024CrossModalConsistency,vanSprang2025SameContent}.
We extend these lines of work by isolating word-level visual styling in text-as-image inputs, allowing us to study its impact independently of textual content.

\noindent{\textbf{Modality gap and representation-level analyses.}}
Contrastive learning for vision and language~\cite{radford2021learning,Jia2021ALIGN,Li2022BLIP} learns a shared embedding space for images and text, but can leave a modality gap between image and text representations even when the image and text describe the same thing~\cite{Liang2022MindTheGap,yamabe2025text}.
Prior work has explored improved alignment objectives and diagnostic metrics that relate internal representation alignment to downstream task behavior~\cite{Eslami2024MitigateGap,Shukor2024ImplicitAlignment}, and has examined the geometric structure of vision and language embedding spaces~\cite{Papadimitriou2025LinearStructure,Bau2017NetworkDissection,Kim2018TCAV}.
Rather than modifying training, we study how semantically preserving changes in color and contrast affect vision encoder representations of VLMs.

\noindent{\textbf{Failures induced by visual prompting.}}
Visual prompting can steer VLMs by adding visual cues, including training-free approaches~\cite{Wu2024VisualPromptSurvey,Wu2024ControlMLLM}.
Previous work has shown that injected or barely visible text can affect VLM behavior, such as in typographic attacks and prompt injection~\cite{OpenAI2021MultimodalNeurons,Azuma2023DefensePrefix,Clusmann2025PromptInjectionOncology}.

Our \emph{Stealth Visual Prompts} vary visual appearance while preserving lexical content, without adding new words or explicit instructions, enabling analysis of visually induced biases.
Prior work on color reliability and shortcut associations motivates our focus on color as a semantically neutral control signal~\cite{Arias2025ColorCLIP,Tang2023CAB}.
While hallucination in VLMs has been widely studied~\cite{Rohrbach2018ObjectHallucination,Liu2024LVLMHallucinationSurvey,Guan2023HallusionBench,Li2023POPE,lee2024volcano}, we show that reduced text contrast can increase hallucinated outputs even when the affected words are task-irrelevant.

\section{Methodology}
\label{sec:method}

We study text understanding in VLMs when text is presented as an image.
Our methodology has three components:
(i) we design \emph{Stealth Visual Prompts} as controlled changes to the visual rendering of text while keeping the underlying string content fixed,
(ii) we construct the \emph{Stealth Prompt Testset} to evaluate end-to-end behavioral changes of VLMs on sentiment analysis and question answering under these prompts, and (iii) we introduce two diagnostic probes (a CLIP representation probe and a VLM-based OCR proxy) to help interpret the observed behavioral effects.

\subsection{\textit{Stealth Visual Prompts}}
\label{sec:method_prompts_rendering}

This section explains how we generate text-as-image as test data and insert \textit{Stealth Visual Prompts} into images. 

\noindent\textbf{Concept definition.}
We define a \textit{Stealth Visual Prompt} as a controlled perturbation applied to the visual rendering of text while keeping the underlying string content fixed.
The goal is to introduce variations that humans typically perceive as ordinary formatting choices, such as emphasis or readability adjustments, rather than explicit instructions.
We focus on \textbf{color} and \textbf{contrast} as they are ubiquitous in real documents and easy to control at the word or span level.
Moreover, the two are complementary: color can encode learned semantic associations, while contrast directly modulates perceptual accessibility.

\noindent\textbf{Text string construction.}
The underlying text strings are generated differently for each subset of the \emph{Stealth Prompt Testset} (Section~\ref{sec:method_datasets}), reflecting the requirements of each task.
For the sentiment sets, we procedurally construct sentences by inserting sentiment-bearing words from a fixed lexicon into neutral templates, ensuring that sentence polarity is controlled by design.
For the VQA set, we sample question--context pairs from SQuAD~\cite{Rajpurkar16} and render the question together with a windowed portion of the corresponding context.

\noindent\textbf{Text-as-image rendering.}
All text strings are rendered onto a standardized $800 \times 600$ pixel canvas with a white background using the DroidSans font with anti-aliasing enabled.
We enable anti-aliasing to match typical document and UI rendering and to avoid aliasing artifacts that could introduce unintended high-frequency cues.
The layout is fixed within each sample, with line breaks and word positions determined once and reused across prompt conditions, and only the targeted visual stylings are modified.

\noindent\textbf{Color prompts.}
Color prompts recolor a predefined subset of words (e.g., sentiment-bearing words), while leaving all other words in black on a white background.
We use six canonical hues (red, green, blue, yellow, cyan, and magenta) and three discrete intensity levels, implemented by fixed RGB channel magnitudes on the active channels of each hue (e.g., red modifies only the R channel, whereas cyan modifies the G and B channels).
Intensity is specified in RGB space rather than matched for perceptual distance; as a result, the perceptual distance to the white background (e.g., measured by $\Delta E$) can vary across hues.
Accordingly, we treat intensity as a rendering-level control parameter.
We implement three intensity levels by setting the RGB channel magnitude to 34 / 85 / 136 (subtle/mild/strong) on the active channels of each hue.

\noindent\textbf{Contrast prompts.}
Contrast prompts reduce text--background contrast by rendering text in low-contrast grayscale on a white background.
Each condition is indexed by a nominal grayscale value for controlled rendering, and we additionally compute a perceptually meaningful contrast measure (CIE $\Delta E$) from the rendered image for analysis.
In the VQA experiments, we consider two variants.
(i) \textit{Global Contrast}, where the entire text is rendered at a given low-contrast level.
(ii) \textit{Saliency Competition}, where one selected span is rendered in high-contrast black while the remaining text is rendered in low contrast, inducing competition for visual saliency.

\begin{figure*}[t]
    \centering
    \includegraphics[width=1.0\textwidth]{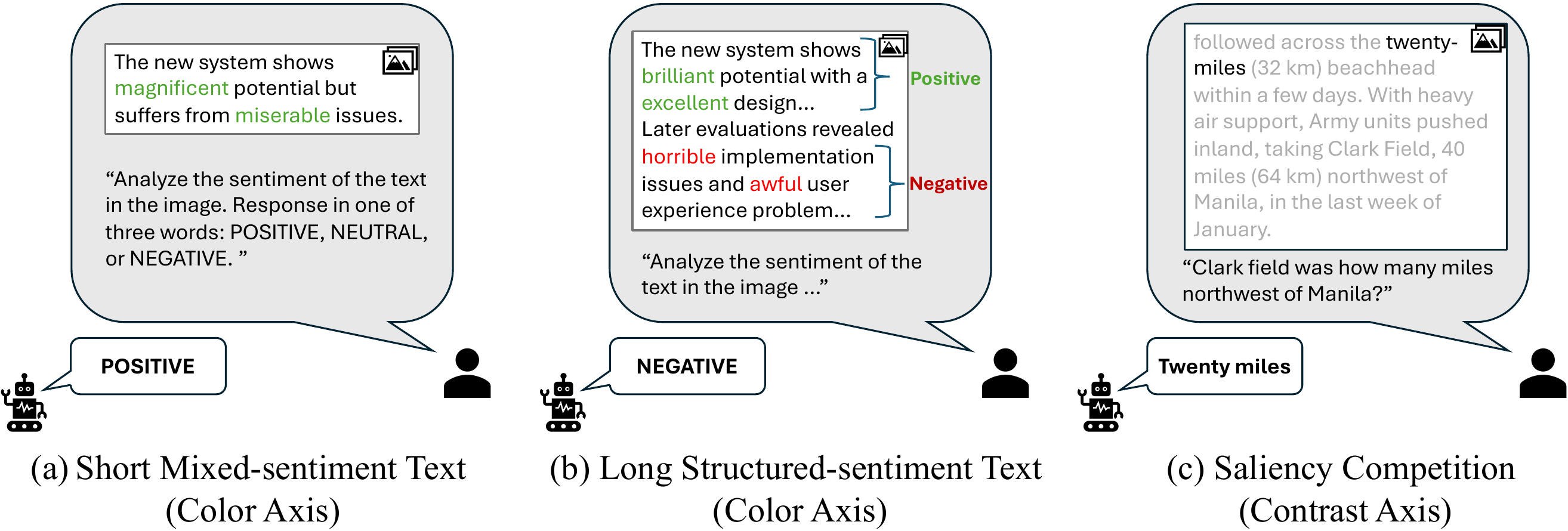} 
    \caption{Examples of the generated visual stimuli. (a) A mixed-sentiment text used in the Color Axis experiment. (b) A structurally separated text. (c) A stimulus from the Saliency Competition (Contrast Axis), where the semantically incorrect decoy (``twenty-miles'') is made visually salient with high contrast.}
    \label{fig:stimuli_examples}
\end{figure*}

\subsection{\emph{Stealth Prompt Testset}}
\label{sec:method_datasets}
We construct the \emph{Stealth Prompt Testset} to probe three distinct behavioral regimes of text-as-image understanding in VLMs, in which the model's dominant strategy can qualitatively change: (i) a \emph{local lexical-integration regime} where all words are clearly visible and the decision should be driven by word-level semantics, (ii) a \emph{structured discourse regime} where longer inputs introduce global structure and positional heuristics (primacy/recency) may dominate, and (iii) an \emph{access-limited regime} where reduced text readability can suppress context utilization and increase reliance on visually salient spans.
Accordingly, the testset comprises three subsets (Figure~\ref{fig:stimuli_examples}):
(a) \textbf{Short-sentence Sentiment Set}, which uses short sentences to isolate word-level color bias;
(b) \textbf{Long-sentence Sentiment Set}, which uses longer, structured sentences to test whether such biases persist or are overridden by discourse structure and positional effects;
(c) \textbf{VQA Stealth Set}, which applies contrast manipulation in VQA to probe failure modes when text accessibility is reduced and visual saliency competes with semantic evidence.

\noindent{\textbf{(a) Short-sentence Sentiment Set}.}
We generate 100 short sentences by injecting sentiment-bearing words from a fixed positive/negative lexicon into templated text with neutral fillers, so that the intended polarity is controlled by design.
For each sentence, we create 37 visual conditions: an all-black baseline and variants where either positive or negative words are colored.
Colors span six canonical hues (red, green, blue, yellow, cyan, magenta) with three predefined intensity levels, yielding one baseline plus $2 \times 6 \times 3$ color conditions.

\noindent{\textbf{(b) Long-sentence Sentiment Set.}}
We generate 100 longer, structured sentences in which positive words are concentrated in the first half and negative words in the second half (or vice versa).
This subset tests whether models adopt positional heuristics (e.g., primacy/recency) under more structured discourse, and whether color-induced bias remains observable in that regime.
We apply the same 37 color conditions as in the short-sentence set.

\noindent{\textbf{(c) VQA Stealth Set.}}
We sample 100 examples from the SQuAD \cite{Rajpurkar16} train split by shuffling once with a fixed seed and selecting the first 100 question--context pairs.
For each example, we window the context around the first ground-truth answer span with a maximum window length of 600 characters (clipped to boundaries),
and render the question and the windowed context as an image.
We evaluate two contrast-based conditions:
\emph{Global Contrast}, where the entire text is rendered at one of six grayscale levels {1, 16, 64, 128, 192, 240} spanning near-black to near-white,
and \emph{Saliency Competition}, where either the ground-truth answer phrase or a decoy word is rendered in high-contrast black, while the remaining text is rendered at one of the six low-contrast levels. Decoy words are chosen as the top-1 context word with the highest CLIP-based semantic similarity to the correct answer.

\subsection{Evaluation Tasks and Metrics}
\label{sec:method_metrics}

\noindent\textbf{Sentiment classification.}
We prompt the model to output one of three labels: \texttt{POSITIVE}, \texttt{NEUTRAL}, or \texttt{NEGATIVE}.
Let $m(\hat{y})\in\{+1,0,-1\}$ map these labels to scalar scores. For visual condition $c$, we define the sentiment bias as
\[
B_c=\frac{1}{N}\sum_{i=1}^{N} m(\hat{y}_{i,c})-\frac{1}{N}\sum_{i=1}^{N}m(\hat{y}_{i,\mathrm{black}}).
\]
Positive values indicate a shift toward \texttt{POSITIVE} predictions relative to the all-black baseline; negative values indicate a shift toward \texttt{NEGATIVE}.

\noindent\textbf{Question answering.}
We report the standard token-level F1 score between the predicted answer and the set of ground-truth answers.
For the Decoy Salient condition, we compute the \emph{Induced Error Rate} as $\mathrm{IER}=\frac{1}{N}\sum_i \mathbf{1}[d_i\subset \hat{a}_i]$, where $d_i$ is the decoy word and $\hat{a}_i$ is the model prediction.
IER is not a general VQA accuracy measure; it directly measures decoy copying under reduced visibility.

\noindent\textbf{Auxiliary diagnostic probes.}
In addition to the Stealth Prompt Testset, we run two controlled single-word probes to interpret the color- and contrast-induced behavioral effects reported in Section~\ref{sec:results}.

\noindent\textbf{CLIP representation probe.}
We probe CLIP \cite{radford2021learning} 
by rendering single words and measuring semantic projections onto ten bipolar axes defined from CLIP text embeddings.
This probe is used to characterize how color styling correlates with systematic shifts in the vision encoder's representation.

\noindent\textbf{VLM-based OCR proxy.}
To calibrate when rendered text becomes effectively inaccessible under reduced contrast, we use each evaluated VLM as a single-word reader with a fixed prompt.
We score each prediction as 1.0 (exact match), 0.5 (substring match), or 0.0 (incorrect) after normalization, and average over the probe vocabulary.
This proxy is intentionally minimal and does not directly model long-context VQA reading; we use it as a diagnostic calibration signal.

\noindent\textbf{Evaluation VLMs.}
We evaluate four open-source VLMs: LLaVA-v1.6-Mistral-7B, LLaVA-v1.6-Vicuna-7B, Qwen2-VL-7B-Instruct, and IDEFICS2-8B. We focus on open-source models to keep the evaluation reproducible under fixed prompts, rendering parameters, and deterministic decoding.

\begin{table*}[t]
    \centering
    \caption{Short-sentence Sentiment Set: worst-case sentiment bias induced by color styling.
    Max Pos. ($\uparrow$) and Max Neg. ($\downarrow$) are measured relative to the all-black baseline.
    \emph{Range} is Max Pos. $-$ Max Neg., summarizing worst-case susceptibility across color conditions.}
    \vspace{-5pt}
    \label{tab:sentiment_bias_spectrum}
    \begin{tabular}{lccc}
        \toprule
        \textbf{Model} & \textbf{Max Pos. $\uparrow$} & \textbf{Max Neg. $\downarrow$} & \textbf{Range} \\
        \midrule
        IDEFICS2-8B         & +0.160 & -0.360 & 0.520 \\
        LLaVA-Mistral-7B    & +0.030 & -0.010 & 0.040 \\
        LLaVA-Vicuna-7B     & +0.060 & -0.060 & 0.120 \\
        Qwen2-VL-7B & +0.420 & -0.480 & 0.900 \\
        \bottomrule
    \end{tabular}
    \vspace{-5pt}
\end{table*}

\section{Experiments and Results}
\label{sec:results}
We evaluate \textit{Stealth Visual Prompts} on the Stealth Prompt Testset.
For each prompt family, we first report end-to-end behavioral effects and then present an auxiliary diagnostic probe to contextualize them: 
a CLIP representation probe for color prompts and a VLM-based OCR proxy for contrast prompts.

\subsection{Color prompts induce systematic sentiment biases}
\label{subsec:results_color}

\noindent\textbf{Short-sentence Sentiment Set (word-level color bias).}
We first evaluate color prompts on short mixed-sentiment sentences, where positive and negative sentiment-bearing words are interspersed.
Table~\ref{tab:sentiment_bias_spectrum} summarizes the maximum sentiment bias over all color conditions, measured relative to the all-black baseline (black text on a white background). For example, \texttt{Qwen2-VL-7B} shows its largest positive bias when positive words are colored green/blue (up to +0.42), and its largest negative bias when negative words are colored red (down to -0.48). Overall susceptibility differs substantially by model: Qwen2-VL-7B shows the largest Total Range (0.90), followed by IDEFICS2-8B (0.52), while the LLaVA variants exhibit much smaller ranges (0.04--0.12), indicating comparatively weaker sensitivity to word-level color styling in this setting.

\begin{figure*}[t]
    \centering
    \includegraphics[width=0.99\linewidth]{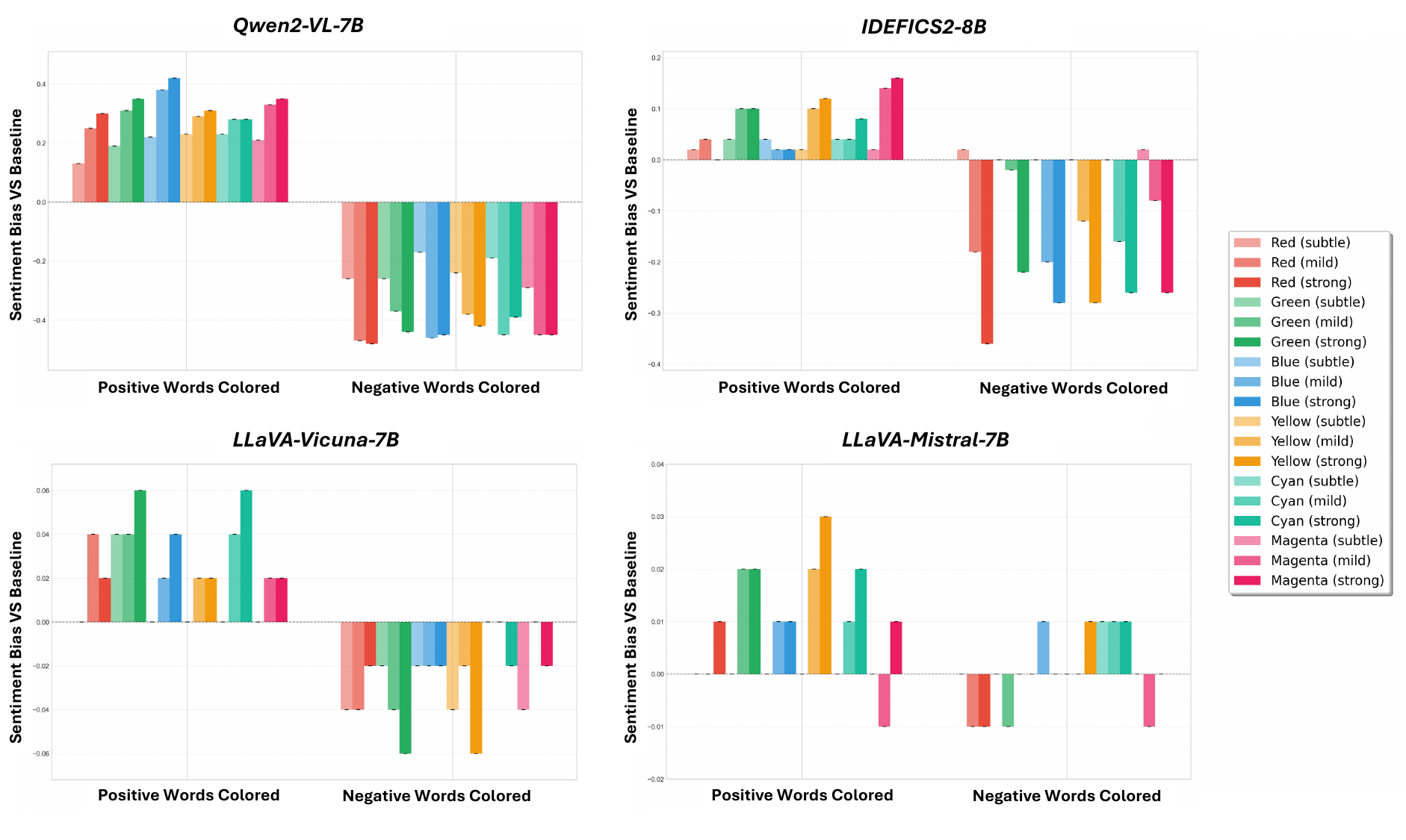}
    \caption{Short-sentence Sentiment Set: average sentiment bias vs.\ baseline for four VLMs. The y-axis is the change in sentiment score induced by the color prompt.}
    \label{fig:sentiment_bias}
\end{figure*}

We observe a clear spectrum of susceptibility: \texttt{Qwen2-VL-7B} exhibits the largest color-induced shifts, while the \texttt{LLaVA} variants are comparatively robust.
Figure~\ref{fig:sentiment_bias} reveals that the bias is not uniform across conditions.
For Qwen2-VL-7B and IDEFICS2-8B, coloring sentiment-bearing words often produces consistent directional shifts, and stronger intensity tends to amplify the magnitude of the bias (dose-response).
In contrast, the LLaVA variants remain close to zero across most hues/intensities, consistent with their small Total Range in Table~\ref{tab:sentiment_bias_spectrum}.

\begin{table*}[t]
    \centering
    \caption{Long-sentence Sentiment Set: positional heuristics under structured discourse.
    \emph{Positional Strategy} indicates whether predictions follow the first half (primacy) or second half (recency).
    \emph{Adherence} is the fraction of samples consistent with that strategy.
    \emph{Color Bias Range} summarizes the residual color-induced variation across color conditions in this structured setting.}
    \vspace{-5pt}
    \label{tab:positional_bias}
    \begin{tabular}{lccc}
        \toprule
        \textbf{Model} & \textbf{Color Bias Range} & \textbf{Positional Strategy} & \textbf{Adherence} \\
        \midrule
        IDEFICS2-8B       & 0.780 & Primacy & 93\% \\
        LLaVA-Mistral-7B  & 0.000 & Recency & 100\%  \\
        LLaVA-Vicuna-7B   & 0.160 & Primacy & 97\% \\
        Qwen2-VL-7B       & 0.000 & Recency & 100\% \\
        \bottomrule
    \end{tabular}
    \vspace{-5pt}
\end{table*}

\noindent\textbf{Long-sentence Sentiment Set (structure and positional heuristics).}
We next evaluate the same color conditions on structured long sentences, where positive and negative words are separated across halves. Rather than treating this as a standard single-label classification problem, we use it to diagnose which part of the sentence dominates the prediction.
We define a model as \emph{primacy} if its predicted polarity matches the first half, and \emph{recency} if it matches the second half; \emph{Adherence} reports how consistently the model follows the identified strategy.
\emph{Color Bias Range} quantifies the residual sensitivity to color styling within this structured regime.

In this regime, models often exhibit a dominant positional heuristic (primacy or recency), and color-induced shifts become secondary.
Table~\ref{tab:positional_bias} summarizes each model's positional strategy and adherence rate, together with the residual \emph{Color Bias Range}.
This indicates that VLMs may switch between simple heuristics depending on discourse structure: color cues can dominate in locally mixed settings, while position can dominate under structured layouts.

\begin{figure*}[t]
    \centering
    \includegraphics[width=0.99\textwidth]{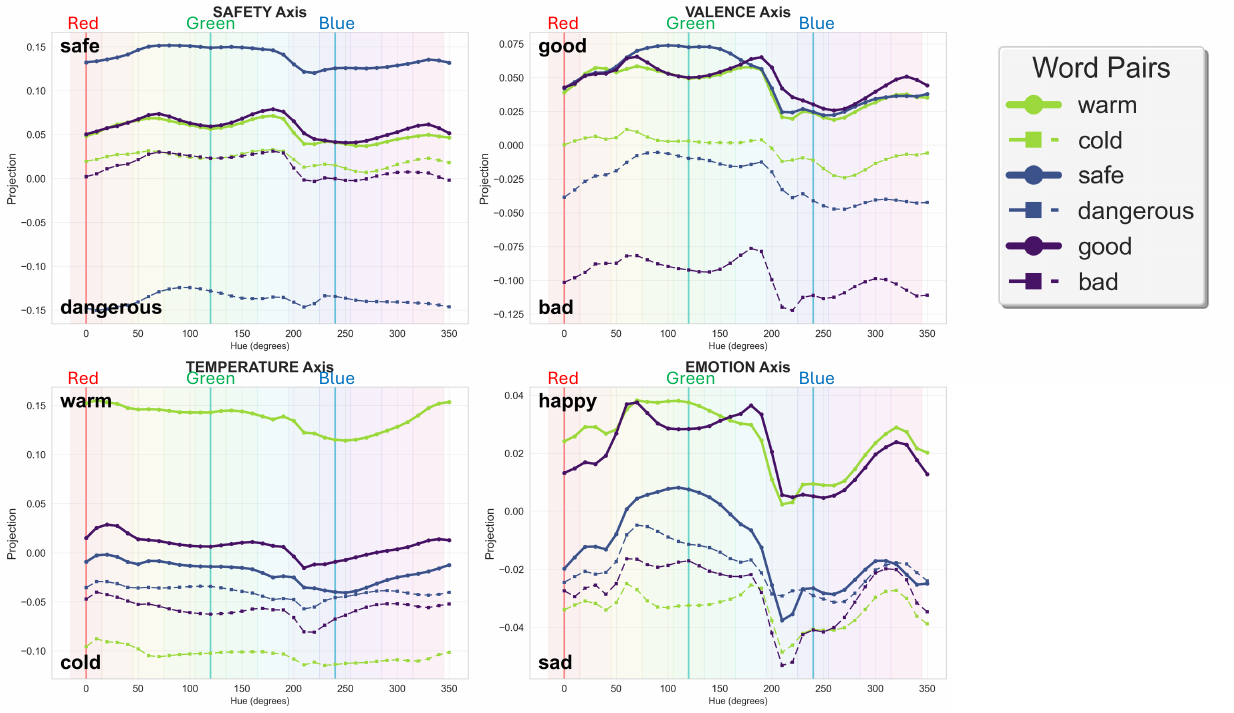}
    \caption{\textbf{CLIP representation probe: hue-dependent semantic projections.}
    We sweep text hue and project CLIP image embeddings of six rendered probe words
    (\texttt{warm}, \texttt{cold}, \texttt{safe}, \texttt{dangerous}, \texttt{good}, \texttt{bad})
    onto four text-defined semantic axes: safety (safe--dangerous), valence (good--bad),
    temperature (warm--cold), and emotion (happy--sad).
    The emotion axis is included as an additional example where hue-dependent shifts are visually apparent; it is defined by the \textit{happy}--\textit{sad} text-embedding difference, while the plotted curves correspond to projections of the same six rendered probe words.}
    \label{fig:hue_tuning_curves}
    \vspace{-10pt}
\end{figure*}

\noindent\textbf{Auxiliary probe: hue systematically shifts vision-encoder semantic projections.}
To interpret why color-based stealth prompts can bias sentiment while keeping the underlying text fixed, we probe vision-encoder representations using the CLIP semantic projection analysis (Section~\ref{sec:method_metrics}).

Figure~\ref{fig:hue_tuning_curves} shows systematic hue-dependent modulation of semantic projections across all four axes.
For example, on the valence axis (top-right), multiple probe words exhibit a shared hue-dependent component: projections are relatively higher around green ($\sim$120$^\circ$) and dip around the blue region ($\sim$250$^\circ$), indicating that hue can shift image embeddings along an abstract semantic direction even when the rendered word itself is unchanged. Valence and emotion show the clearest modulation in our visualization, while safety and temperature exhibit smaller but still systematic variation.

While CLIP is used here as a diagnostic encoder and does not imply a literal mechanism for all evaluated architectures, the observed representation shifts are consistent with the color-induced sentiment biases above.

\subsection{Contrast prompts increase saliency-driven errors in VQA}
\label{subsec:results_contrast}
We evaluate contrast prompts on the VQA Stealth Set using \textit{Global Contrast} and \textit{Saliency Competition} (Section~\ref{sec:method_datasets}).
We report (i) token-level F1 as a task-success metric and (ii) the \emph{Induced Error Rate}, defined in the \textit{Decoy Salient} condition as the fraction of predictions that contain the decoy word.
We use the induced error as an operational indicator of saliency-driven failures, especially for models whose absolute F1 is near zero.

\begin{table*}[t]
    \centering
    \caption{VQA Stealth Set: average F1 in Saliency Competition, aggregated over the six low-contrast grayscale levels. Columns differ only in which span is rendered in high-contrast black (none / answer / decoy).}
    \vspace{-5pt}
    \label{tab:saliency_f1_comparison}
    \begin{tabular}{lccc}
        \toprule
        \textbf{Model} & \textbf{Baseline F1} & \textbf{Answer Salient F1} & \textbf{Decoy Salient F1} \\
        \midrule
        IDEFICS2-8B       & 0.054 & 0.064 (+18.5\%) & 0.052 (-3.7\%) \\
        LLaVA-Mistral-7B  & 0.053 & 0.058 (+9.4\%)  & 0.052 (-1.9\%) \\
        LLaVA-Vicuna-7B   & 0.056 & 0.061 (+8.9\%)  & 0.055 (-1.8\%) \\
        Qwen2-VL-7B & 0.745 & 0.828 (+11.1\%) & 0.738 (-0.9\%) \\
        \bottomrule
    \end{tabular}
\end{table*}

\begin{table*}[t]
    \centering
    \caption{VQA Stealth Set: Induced Error Rate (\%) in the \textit{Decoy Salient} condition at the six low-contrast grayscale levels used in our sweep (higher value $\Rightarrow$ lower visibility).}

    \vspace{-5pt}
    \label{tab:saliency_induced_error}
    \begin{tabular}{lcccccc}
        \toprule
        \textbf{Model} & \textbf{1} & \textbf{16} & \textbf{64} & \textbf{128} & \textbf{192} & \textbf{240} \\
        \midrule
        IDEFICS2-8B       & 24\% & 23\% & 24\% & 27\% & 32\% & 36\% \\
        LLaVA-Mistral-7B  & 27\% & 25\% & 26\% & 24\% & 24\% & 27\% \\
        LLaVA-Vicuna-7B   & 19\% & 19\% & 20\% & 20\% & 22\% & 25\% \\
        \midrule
        Qwen2-VL-7B & 4\% & 4\% & 4\% & 4\% & 5\% & 6\% \\
        \bottomrule
    \end{tabular}
\end{table*}

\begin{wrapfigure}{r}{0.5\textwidth}
    \centering
    \includegraphics[width=0.48\textwidth]{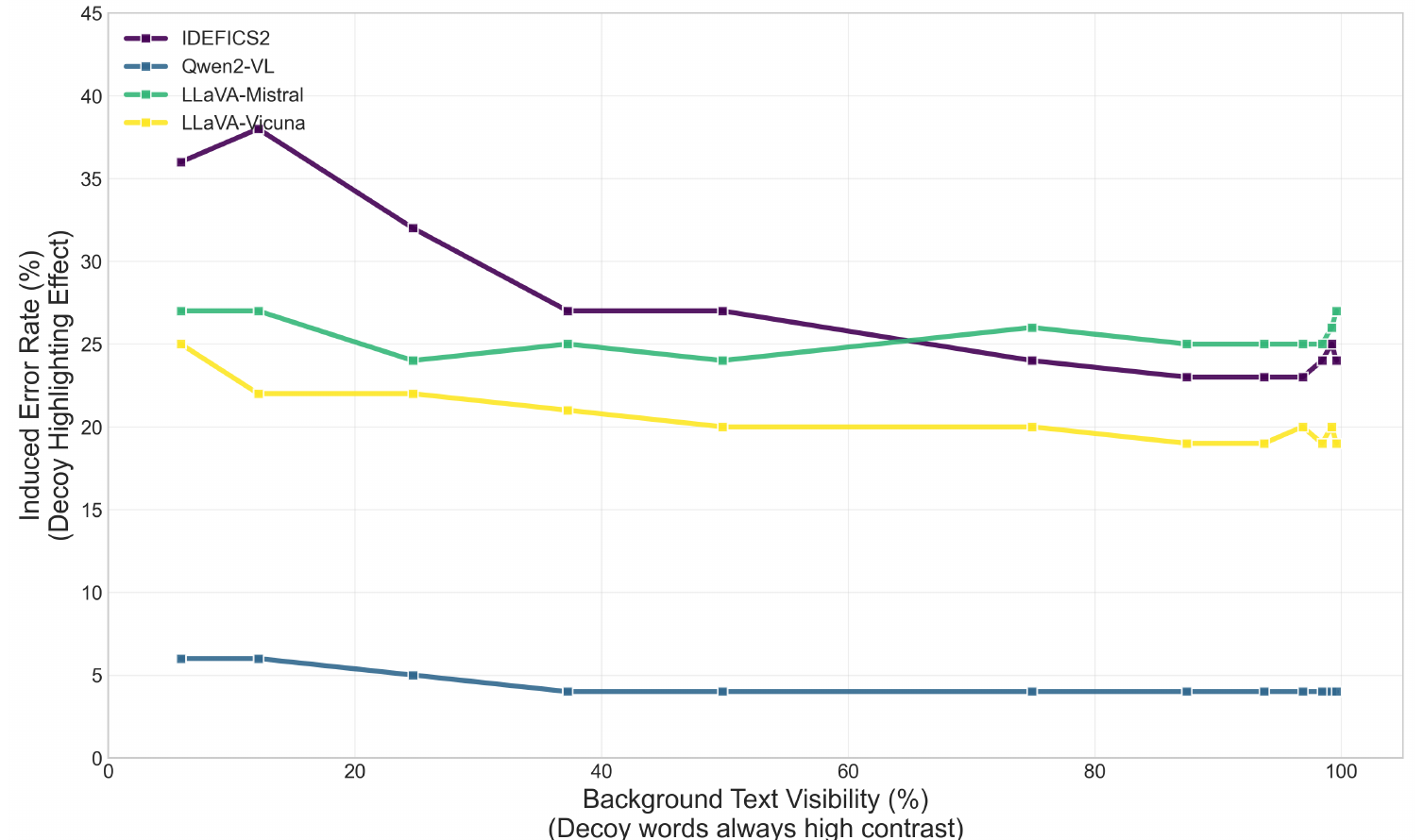}
    \caption{VQA Stealth Set: Induced Error Rate as a function of the grayscale level used for the non-salient context text (higher $\Rightarrow$ closer to white and lower contrast).}
    \label{fig:saliency_competition}
\end{wrapfigure}

\noindent\textbf{Saliency Competition under reduced visibility.}
Table~\ref{tab:saliency_f1_comparison} shows a clear capability gap: \texttt{Qwen2-VL-7B} benefits substantially when the answer span is made salient, whereas \texttt{IDEFICS2-8B} and the \texttt{LLaVA} variants remain at very low F1 with only marginal changes across columns, indicating limited end-to-end reading in this setup.
We therefore focus on the \emph{Induced Error Rate (IER)} as a direct indicator of saliency-driven decoy copying in the \textit{Decoy Salient} condition.
Table~\ref{tab:saliency_induced_error} and Fig.~\ref{fig:saliency_competition} show that as the non-salient context text becomes less visible (higher grayscale value), IER increases for \texttt{IDEFICS2-8B} from 24\% to 36\% and for \texttt{LLaVA-Vicuna-7B} from 19\% to 25\%, while \texttt{LLaVA-Mistral-7B} stays high and relatively flat (24--27\%) and \texttt{Qwen2-VL-7B} remains low at 4--6\%.
Overall, these results indicate that contrast reduction can shift some VLMs toward saliency-driven shortcut behavior even though the lexical content is unchanged.

\setlength{\intextsep}{3pt}
In contrast, \texttt{Qwen2-VL-7B} stays low (4--6\%), suggesting stronger robustness to visually salient but incorrect cues.

\begin{figure}[t]
    \centering
    \includegraphics[width=0.99\linewidth]{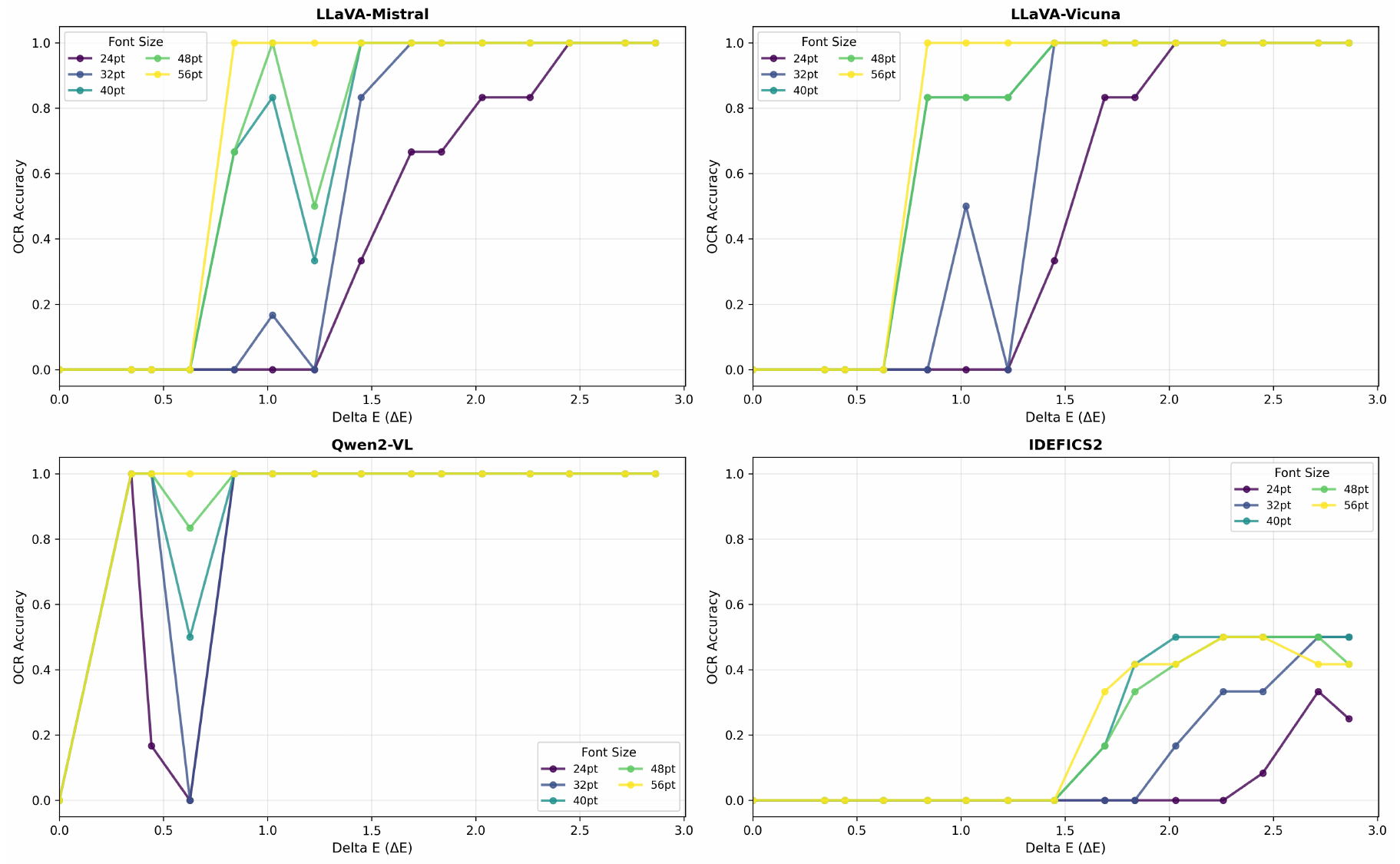}
    \caption{Model-dependent readability transitions measured by a VLM-based OCR proxy. OCR accuracy is plotted as a function of text--background contrast ($\Delta E$) for multiple font sizes. The transition from low to high accuracy occurs over a relatively narrow range and varies by model and font size.}
    \label{fig:perceptual_cliff}
\end{figure}

\noindent\textbf{OCR proxy.}
To contextualize the contrast sweep, we additionally measure a minimal OCR proxy (Section~\ref{sec:method_metrics}) using single-word stimuli.
Figure~\ref{fig:perceptual_cliff} shows a non-linear \emph{readability transition}: OCR accuracy rises sharply over a relatively narrow contrast range, and the transition location depends on both model and font size.
This probe does not directly model long-context VQA reading, but it provides a calibration signal that small contrast changes can move a model from a low-readability to a high-readability region for rendered text under our prompting/setup.
This is consistent with the increased saliency-driven errors we observe when the non-salient context text becomes less accessible.

\section{Discussion}
Our results show that VLMs are sensitive to the visual form of rendered text: semantically identical strings can yield different outputs under ordinary formatting. Color can act as an implicit control channel---recoloring a small subset of sentiment-bearing words biases sentiment predictions, and a CLIP-based probe reveals that sweeping hue shifts vision-encoder image embeddings along human-interpretable semantic axes (e.g., valence) even when the rendered word is fixed.

Contrast primarily affects access to contextual evidence. As text--background contrast decreases, several models rely more on visually salient spans, which increases decoy-driven errors in VQA\@. A minimal single-word OCR proxy exhibits a model-dependent \emph{readability transition}, consistent with the idea that small contrast changes can move models between lower- and higher-access regimes for visual text under our setup, making saliency cues more influential.

These sensitivities imply a reliability and safety risk for VLM pipelines that ingest documents or UI screenshots: benign or adversarial styling can steer model decisions without changing the underlying text. Practical safeguards include normalizing rendered text before inference, cross-checking image-based answers with OCR-extracted text, and adding style-invariance checks to evaluation suites. Limitations include English-only stimuli, RGB-defined intensity levels, and the OCR proxy's limited scope; future work should test broader rendering factors such as fonts, layout, and multilingual scripts.

\section{Conclusion}

We introduced \textit{Stealth Visual Prompts}---semantics-preserving changes to the visual rendering of text---as a controlled methodology for studying rendered-text understanding in VLMs.
The Stealth Prompt Testset shows that word-level color styling can systematically bias sentiment predictions and sometimes override contradictory lexical evidence; under structured long sentences, models often shift to positional heuristics.
For contrast, reducing text--background visibility increases reliance on visually salient shortcut cues and can induce decoy-driven VQA errors.

The auxiliary probes contextualize these effects: CLIP embeddings shift with hue along semantic axes, and the OCR proxy shows model-dependent \emph{readability transitions}.
Overall, ordinary formatting should not be treated as purely cosmetic for text-as-image inputs.
It can act as an implicit control channel, motivating evaluation protocols and robustness methods that explicitly account for visual styling in deployed VLM pipelines.

\section*{Acknowledgments}
This work was supported by the AIST policy-based budget project ``R\&D on Generative AI Foundation Models for the Physical Domain''. We used ABCI 3.0 provided by AIST and AIST Solutions with support from ``ABCI 3.0 Development Acceleration Use''.

\bibliographystyle{splncs04}
\bibliography{main}

\end{document}